\documentclass[lettersize,journal]{IEEEtran}
\usepackage{amsmath,amsfonts}
\usepackage{algorithmic}
\usepackage{algorithm}
\usepackage{array}
\usepackage[caption=false,font=normalsize,labelfont=sf,textfont=sf]{subfig}
\usepackage{textcomp}
\usepackage{stfloats}
\usepackage{url}
\usepackage{verbatim}
\usepackage{graphicx}

\usepackage{cite}
\usepackage{diagbox}
\usepackage{multirow} 
\usepackage{booktabs}
\usepackage{bbding}

\usepackage[table]{xcolor}
\begin{document}
	
	\title{Semantic Manipulation Localization}
	
	\author{Zhenshan Tan, Chenhan Lu, Yuxiang Huang, Ziwen He, Xiang Zhang, Yuzhe Sha, Xianyi Chen, Tianrun Chen, Zhangjie Fu 
		\thanks{Zhenshan Tan, Chenhan Lu, Yuxiang Huang, Xiang Zhang, Ziwen He, Yuzhe Sha, Xianyi Chen, and Zhangjie Fu are with the Engineering Research Center of Digital Forensics, Ministry of Education, Nanjing University of Information Science and Technology, Nanjing, 210044, China (e-mail: zstan@nuist.edu.cn).}
		\thanks{Tianrun Chen is with the College of Computer Science and Technology, Zhejiang University, and KOKONI3D, Moxin (Hangzhou) Technology Company Ltd., Hangzhou 310027, China (e-mail: tianrun.chen@zju.edu.cn)}
	}
	
	\markboth{Journal of \LaTeX\ Class Files,~Vol.~14, No.~8, August~2021}%
	{Shell \MakeLowercase{\textit{et al.}}: A Sample Article Using IEEEtran.cls for IEEE Journals}
	
	
	\maketitle
	
	\begin{abstract}
		Image Manipulation Localization (IML) aims to identify edited regions in an image. However, with the increasing use of modern image editing and generative models, many manipulations no longer exhibit obvious low-level artifacts. Instead, they often involve subtle but meaning-altering edits to an object’s attributes, state, or relationships, while remaining highly consistent with the surrounding content. This makes conventional IML methods less effective, as they mainly rely on artifact detection rather than semantic sensitivity. To address this issue, we introduce Semantic Manipulation Localization (SML), a new task that focuses on localizing subtle semantic edits that significantly change image interpretation. We further construct a dedicated fine-grained benchmark for SML using a semantics-driven manipulation pipeline with pixel-level annotations. Based on this task, we propose TRACE (Targeted Reasoning of Attributed Cognitive Edits), an end-to-end framework that models semantic sensitivity through three progressively coupled components: semantic anchoring, semantic perturbation sensing, and semantic-constrained reasoning. Specifically, TRACE first identifies semantically meaningful regions that support image understanding, then injects perturbation-sensitive frequency cues to capture subtle edits under strong visual consistency, and finally verifies candidate regions through joint reasoning over semantic content and semantic scope. Extensive experiments show that TRACE consistently outperforms existing IML methods on our benchmark and produces more complete, compact, and semantically coherent localization results. These results demonstrate the necessity of moving beyond artifact-based localization and provide a new direction for image forensics in complex semantic editing scenarios.
	\end{abstract}
	
	
	\section{Introduction}
	\label{sec:intro}
	Image Manipulation Localization (IML) targets the identification of edited or tampered regions in an image and serves as a fundamental task in computer vision. Recent methods have achieved strong performance on public benchmarks by detecting low-level cues, including statistical irregularities ~\cite{lou2025exploring, huang2025forensic}, texture inconsistencies~\cite{zhuang2023reloc}, and boundary artifacts~\cite{nam2025m2sformer}.
	
\begin{figure}[t]
		\centering
		\includegraphics[width=0.5\textwidth]{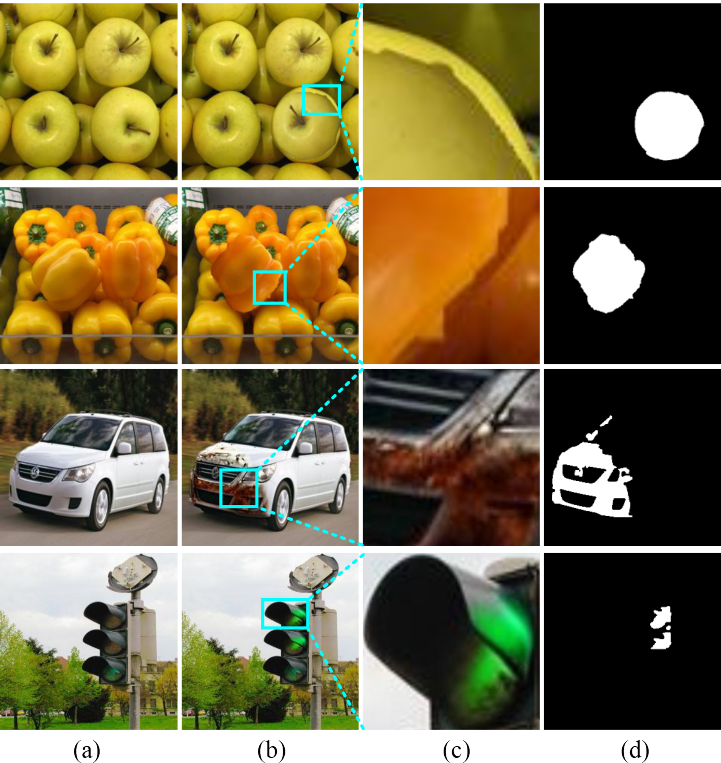}
		\caption{Distinction between image manipulation localization (IML) and semantic manipulation localization (SML). (a) Authentic images; (b) Manipulated images; (c) Local enlarged areas of manipulated images; (d) Ground-truth.}
		\label{fig1}
	\end{figure}

	However, we observe that many manipulations do not exhibit clear low-level anomalies. Instead, they often perform high-level semantic edits to an object’s attributes, state, or relationships. The edited region can be highly localized, yet it is sufficient to substantially change the overall semantic interpretation of the image. Fig.~\ref{fig1} illustrates the distinction between these two types of manipulations. The first two rows show classic IML examples ~\cite{wen2016coverage}, where detection mainly targets traces and distribution shifts, while the underlying semantics may remain unchanged. The last two rows show semantic manipulations that modify key regions to significantly alter the intended meaning of the image, making it misleading or negative. Notably, the modified regions can be very small. Moreover, advanced editing techniques combined with such small edits make traces and distribution shifts difficult to detect. As a result, existing IML paradigms are limited when facing this type of manipulation.
	
	To explain this gap, we analyze how semantic edits influence image understanding. Semantics are not perceived uniformly across the image. Both humans and vision models tend to rely on a small set of semantically decisive regions, which are often salient~\cite{liu2019simple,tan2021depth}. Subtle yet semantically consistent edits within these regions, such as changing a key attribute or state, can induce a large semantic shift while preserving overall appearance. As a result, semantic manipulations are often embedded in high-semantic-load regions and remain highly consistent with surrounding content, weakening low-level cues that traditional methods depend on. Moreover, semantic importance varies within a salient region, making the choice of where to edit more critical than how much to edit, and raising the bar for localization.
	
	Motivated by these observations, we introduce and study Semantic Manipulation Localization (SML). Unlike IML~\cite{dong2022mvss,niloy2023cfl,liu2023making, han2024hdf,du2025forensichub,wang2025boosting} that relies on low-level anomalies, SML centers on semantic sensitivity and aims to localize semantic edits under strong visual consistency. This shifts the goal of localization away from texture irregularities or boundary artifacts and toward a more fundamental question: which changes even subtle changes can look visually unchanged yet trigger a different global semantic interpretation. The key to solving SML is to turn semantic sensitivity from an abstract notion into a computable and localizable criterion. A model should identify the semantic scope that carries meaning, capture semantic perturbations that alter meaning despite appearance consistency, and locate sematic manipulation areas.
	To address SML task, we build a dataset tailored for SML using a semantics-driven manipulation pipeline and provide pixel-level annotations for fine-grained local edits. Further, we propose TRACE (Targeted Reasoning of Attributed Cognitive Edits), which formulates SML as semantic sensitivity modeling rather than anomaly detection. TRACE consists of three tightly coupled components, including a semantic anchoring module that grounds high-level interpretation to the key objects and their spatial scope, a semantic perturbation sensing module that extracts fine-grained cues that can trigger semantic shifts under visual consistency, and a semantic-constrained reasoning module that verifies and consolidates candidate regions under global semantic logic to predict stable and interpretable results. TRACE maps an input image to a pixel-level manipulation mask in an end-to-end manner and progressively refines predictions across semantic levels toward the final localization.
	
	Our contributions can be summarized as follows. 
	
	\begin{itemize}
		\item We introduce Semantic Manipulation Localization, a new task setting that targets the localization of meaning-altering semantic edits.
		\item We construct a fine-grained SML dataset using a semantics-driven manipulation pipeline, providing pixel-level annotations for localized semantic edits within salient targets to support systematic training and evaluation.
		\item We propose TRACE, an end-to-end framework that models semantic sensitivity by combining semantic anchoring, semantic perturbation sensing, and semantic-constrained reasoning to locate manipulation areas.
	\end{itemize}
	
	\section{Related Works}
	\label{sec:rel}
	In this work, we address semantic manipulation localization (SML), a task that fundamentally differs from conventional IML by emphasizing semantic sensitivity rather than low-level artifact detection. This problem is closely related to two research directions: 1) image manipulation localization, which provides the foundational framework for identifying manipulated regions, and 2) high-density semantic information modeling, which focuses on identifying regions that carry concentrated semantic meaning and play a decisive role in image understanding. Accordingly, we review related work from these two perspectives.
	
	\subsection{Image Manipulation Localization}
	IML aims to identify tampered regions within images. In recent years, as manipulation techniques have become increasingly diverse and imperceptible, research has gradually shifted from modeling single forensic clues to developing methods with stronger generalization ability and robustness. Dong \emph{et al.}~\cite{dong2022mvss} proposed a multi-view multi-scale supervised framework, which introduces supervisory signals at different views and scales to effectively enhance robustness and specificity under cross-dataset conditions. Niloy \emph{et al.}~\cite{niloy2023cfl} proposed a contrastive learning-based approach that constructs positive and negative feature pairs to explicitly separate the feature distributions of untampered and manipulated regions, thereby improving cross-domain generalization without relying on specific manipulation traces.
	
	To better exploit low-level forensic clues, Han \emph{et al.}~\cite{han2024hdf} proposed a dual-stream network that fuses RGB and SRM features, and designed a three-level complementary module to progressively enhance the modeling of structural discrepancies and boundary details in tampered regions. Building on the complementary nature of different artifacts and traces, Triaridis \emph{et al.}~\cite{triaridis2023mmfusion} proposed a two-stage framework that integrates RGB images with responses from multiple noise-sensitive filters, allowing joint modeling of image manipulation localization and detection. By incorporating an adaptive feature re-weighting mechanism, their framework achieves competitive performance.
	
	With the growing demand for global modeling, Ma \emph{et al.}~\cite{ma2023iml} identified the limitations of CNNs in long-range dependency modeling and non-semantic artifact extraction, and proposed an end-to-end ViT-based image manipulation localization framework with high-resolution representations, multi-scale supervision, and explicit manipulation boundary constraints, which enhances the characterization of non-semantic manipulation artifacts and regional discrepancies. To overcome the insufficient cross-modal interaction in traditional RGB–noise feature concatenation, Guo \emph{et al.}~\cite{guo2024effective} proposed an enhanced Transformer encoder with dual branches for image manipulation localization, introducing a cross-scale collaborative attention mechanism to enable deep interaction and effective complementarity among multi-source features.
	
	Generative image manipulation detection has emerged as a new direction in image forensics. The core challenge lies in the absence of obvious pixel-level artifacts in generative manipulated images, with only subtle semantic or texture anomalies present. Du \emph{et al.}~\cite{du2025forensichub} introduced a unified benchmark and codebase for all-domain fake image detection and localization, which supports unified training and testing of multiple forensic models across tasks. 
	
	In general, recent image manipulation localization methods exhibit several notable trends: a shift from single-clue modeling to multi-clue collaborative localization \cite{guillaro2023trufor,nam2025m2sformer}, a transition from strong supervision to generalization-oriented learning paradigms \cite{wu2025rethinking}, and the adoption of more powerful global backbone networks and efficient modeling mechanisms \cite{kwon2025safire}.
	
	\subsection{High-Density Semantic Information Modeling}
	Salient object detection (SOD) can be regarded as a practical and widely adopted proxy for modeling high-density semantic regions \cite{tan2024co,tan2023bridging,chen2022utc,tan2024bridging}. Therefore, in this work, we directly employ SOD as an implementation for high-density semantic region modeling. The SOD aims to identify the most visually attractive regions in an image by simulating the human visual system. In recent years, research on salient object detection has mainly focused on multi-scale feature fusion, edge detail enhancement, and global context modeling. Early SOD methods were predominantly based on CNNs, focusing on efficient feature aggregation and boundary preservation. Liu \emph{et al.}~\cite{liu2019simple} proposed an efficient salient object prediction approach using a simple pooling-based feature fusion strategy. Zhao \emph{et al.}~\cite{zhao2019egnet} introduced an edge-guided module that leverages edge information to assist in predicting the salient region, demonstrating superior boundary preservation ability and segmentation accuracy compared with conventional CNN-based networks.
	
	Recently, more powerful backbone networks and their corresponding optimization strategies have become the mainstream research direction. With the successful application of Transformers in vision tasks, researchers have begun to introduce self-attention mechanisms into salient object detection. Zeng \emph{et al.}~\cite{zeng2023dual} proposed a dual swin-transformer based mutual interactive network capable of learning contextualized, dense, and edge-aware features, and strengthening cross-level feature interaction through self-attention mechanisms. 
	
	Foundation vision models have also been introduced into SOD due to their strong generalization ability. Xiong \emph{et al.}~\cite{xiong2026sam2} integrated the hierarchical Hiera backbone of SAM2 into a U-shaped encoder–decoder architecture to fully leverage its representation capability and generalization performance in segmentation tasks. By incorporating lightweight adapters for efficient fine-tuning, the framework effectively utilizes the hierarchical features from the SAM2 encoder, thereby improving performance across various downstream tasks, such as salient object detection. He \emph{et al.}~\cite{he2025samba} proposed a unified salient object detection framework based entirely on the Mamba architecture, which significantly reduces computational complexity while preserving global sequence modeling capability, providing a new efficient modeling paradigm for salient object detection.
	
	Overall, modern salient object detection methods are evolving toward a stronger global semantic modeling capability \cite{ren2024unifying}, a more refined local structural perception \cite{yuan2025uncertainty}, and a more efficient feature extraction mechanism \cite{li2024conda}, aiming to accurately characterize the most attention-worthy regions in complex scenes.
	
	\section{Dataset}\label{sec3}
	As there is currently no dedicated dataset for semantic tampering localization, we construct a broad and effective benchmark ourselves. Our dataset is built through two complementary strategies. The first is a two-stage automatic semantically decisive region identification and semantic tampering pipeline. The second is a set of selected and synthetic cognition-relevant samples.
	
	\begin{figure}[t]
		\centering
		\includegraphics[width=0.48\textwidth]{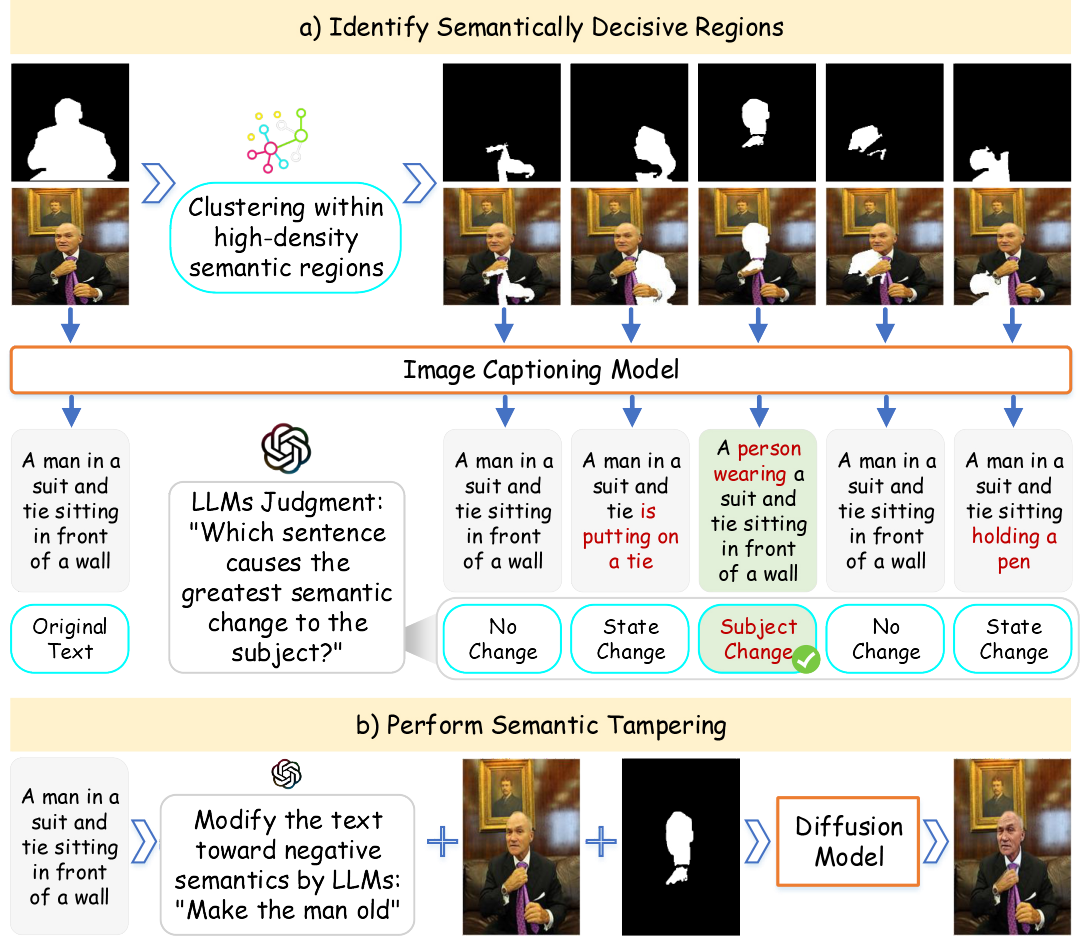}
		\caption{The pipeline of two-stage automatic semantically decisive region identification and semantic tampering.}
		\label{fig2}
	\end{figure}
	
	\subsection{Semantically Decisive Region Identification and Semantic Tampering}\label{sec3.1}
	This pipeline is shown in Fig.~\ref{fig2}. The pipeline first identifies local regions that are decisive for understanding the subject semantics in an image, and then performs semantic tampering. This process produces manipulated samples with clear semantic attack targets and localization annotations.
	
	The key idea is that different image regions do not contribute equally to overall semantic understanding. Compared with large background areas that have limited semantic value, local high-information regions inside salient objects often contain key cues about the subject category, attributes, or state~\cite{tan2022co}. Even a small change in such regions may strongly affect how a viewer understands the image subject. Therefore, dataset construction should not rely on randomly selected editing regions. Instead, it should focus on local regions that are most critical to subject-level semantic interpretation.
	
	However, it is difficult to directly determine which region is the most important from pixels alone. In contrast, it is more practical to estimate region importance through the change in textual descriptions after local masking. Based on this idea, we reformulate key region identification as an indirect process of local masking, caption generation, and semantic comparison. Specifically, we first extract the salient object region from the image. We then apply K-means clustering with spatial distance constraints inside this region to obtain several candidate local regions. To improve spatial coherence and structural quality, we further refine the clustering results with morphological operations, including opening, closing, dilation, and erosion. The number of clusters is set to 5 by default and is manually adjusted between 3 and 5 according to the target structure. Next, each candidate region is used as a mask on the original image to generate a set of locally masked images. The original image and the masked images are then fed into an image captioning model ~\cite{mokady2021clipcap} to obtain their corresponding descriptions. An LLM such as ChatGPT is used to compare the original text with each masked-image text. This comparison is not based on the amount of text change. Instead, we use a subject-oriented hierarchical criterion. We first examine whether local masking changes the identification of the image subject itself. If the subject remains unchanged, we then examine whether the masking affects the understanding of the subject’s state, attributes, or actions. Candidate regions that cause larger subject-level semantic changes are treated as more semantically decisive. Finally, we manually review the selected results to ensure the quality of region selection.
	
	After obtaining the semantically decisive regions, we move to the semantic tampering stage. Specifically, based on the caption of the original image, we use an LLM such as ChatGPT to generate a negatively modified semantic instruction that stays aligned with the original meaning but introduces a targeted semantic shift, such as changing the subject’s attributes, category, or state. The original image, the key-region mask, and the semantic editing instruction are then fed into an image editing model ~\cite{brooks2023instructpix2pix}. Under spatial constraints, the model performs local semantic tampering on the key region, producing a manipulated image and its corresponding localization annotation. Finally, we manually remove failed edits, samples with weak semantic changes, and samples with strong artifacts, yielding high-quality data for semantic tampering localization.
	
	It is important to note that our method does not require candidate regions to perfectly align with complete semantic units. Instead, local regions with moderate noise and imperfect boundaries better match real attack scenarios, since attackers often cannot edit a local semantic unit in a fully precise and ideal way. Such imperfect regions can also reduce overfitting to regular boundaries and encourage the model to focus on more subtle and fine-grained semantic tampering cues. Since excessive noise may increase learning difficulty, we also construct an additional setting with relatively complete semantic tampering as a complementary scheme, which is shown in subsection~\ref{sec3.2}. In this part, we produce 4899 images and corresponding masks.
	
	\subsection{Selected and Synthetic Cognition-Relevant Samples}\label{sec3.2}
	We further construct a dataset partition termed Selected and Synthetic Cognition-Relevant Samples from public manipulation datasets and public saliency datasets. This partition is built through two routes: selecting cognition-relevant samples from existing manipulation datasets and synthesizing new forgery samples from saliency datasets. Unlike subsection~\ref{sec3.1}, which is based on semantically decisive region identification and targeted semantic tampering, this partition is built through public-data selection and salient-object-based forgery generation.
	
	For public manipulation datasets, we extract salient regions using a pre-trained saliency detection network~\cite{liu2019simple} and measure their overlap with the ground-truth manipulation masks. Images with more than 80\% overlap are kept as cognition-relevant manipulation samples, since their manipulated regions mainly lie in perceptually important areas.
	
	For public saliency datasets, we synthesize forged samples. We consider three representative manipulation types: splicing, copy-move, and removal. For splicing and copy-move, salient regions are extracted as donor objects, randomly transformed, and then either pasted into another image or relocated within the same image. For removal, we employ an inpainting algorithm~\cite{li2022mat} and a text-guided inpainting model based on Stable Diffusion~\cite{rombach2022high} to generate semantically consistent or semantically altered salient-region forgeries. All selected and synthesized samples are manually reviewed and refined to ensure data quality. The detailed composition is summarized in Table~\ref{tab1}, which includes 18,733 images and corresponding masks. 
	
	In total, our dataset contains 23,632 pairs of authentic images, manipulated images, and corresponding masks. We initially split the two subsets into training, validation, and test sets with a ratio of 8:1:1. However, we observe that the test set included too few challenging samples from Semantically Decisive Region Identification and Semantic Tampering. We therefore move a portion of hard samples from the training set to the test set. The final split consists of 17,282 training pairs, 2,860 validation pairs, and 3,490 test pairs.
	
	\begin{table}[t]
		\caption{Dataset composition source. `S' means saliency and `M' means manipulation. `C' represents category. `SP' denotes splicing, `CM' denotes copy-move, and `RM' denotes removal.}
		\label{tab1}
		\centering
		\resizebox{\columnwidth}{!}{%
			\begin{tabular}{cccccccc}
				\hline
				\multirow{2}{*}{Datasets} & \multirow{2}{*}{C} & \multicolumn{3}{c}{Tampering Type} & \multirow{2}{*}{Size} & \multirow{2}{*}{Total} & \multirow{2}{*}{Selected} \\
				\cline{3-5}
				& & SP & CM & RM & & & \\
				\hline
				CASIA~\cite{dong2013casia} & M & \checkmark & \checkmark & & 410×309 & 6041 & 734 \\
				CMFD~\cite{cozzolino2014copy} & M & & \checkmark & & 1024×768 & 80 & 2 \\
				COVERAGE~\cite{wen2016coverage} & M & \checkmark & \checkmark & & 469×403 & 100 & 53 \\
				data-tifs-2016-map~\cite{korus2016multi} & M & \checkmark & & & 480×270 & 136 & 9 \\
				IMD2020~\cite{novozamsky2020imd2020} & M & \checkmark & & \checkmark & 1040×876 & 3781 & 914 \\
				realistic-tampering-dataset~\cite{korus2016multi} & M & \checkmark & & \checkmark & 1920×1080 & 220 & 16 \\
				CoSal2015~\cite{zhang2016detection} & S & \checkmark & \checkmark & & 465×410 & 2015 & 879 \\
				DUTS-TE~\cite{wang2017learning} & S & & \checkmark & & 382×327 & 5019 & 598 \\
				DUTS-TR~\cite{wang2017learning} & S & & \checkmark & & 380×326 & 10553 & 1000 \\
				ECSSD~\cite{shi2015hierarchical} & S & & \checkmark & & 374×317 & 1000 & 387 \\
				HKU~\cite{li2015visual} & S & \checkmark & \checkmark & \checkmark & 385×290 & 6947 & 5557 \\
				SOD~\cite{movahedi2010design} & S & & & & 436×362 & 300 & 69 \\
				DUTOMRON~\cite{yang2013saliency} & S & \checkmark & \checkmark & & 375×320 & 5166 & 1807 \\
				MSRA-B~\cite{jiang2013salient} & S & \checkmark & & & 373×319 & 7500 & 6708 \\
				\hline
			\end{tabular}
		}
	\end{table}
	
	\begin{figure*}[t]
		\centering
		\includegraphics[width=\textwidth]{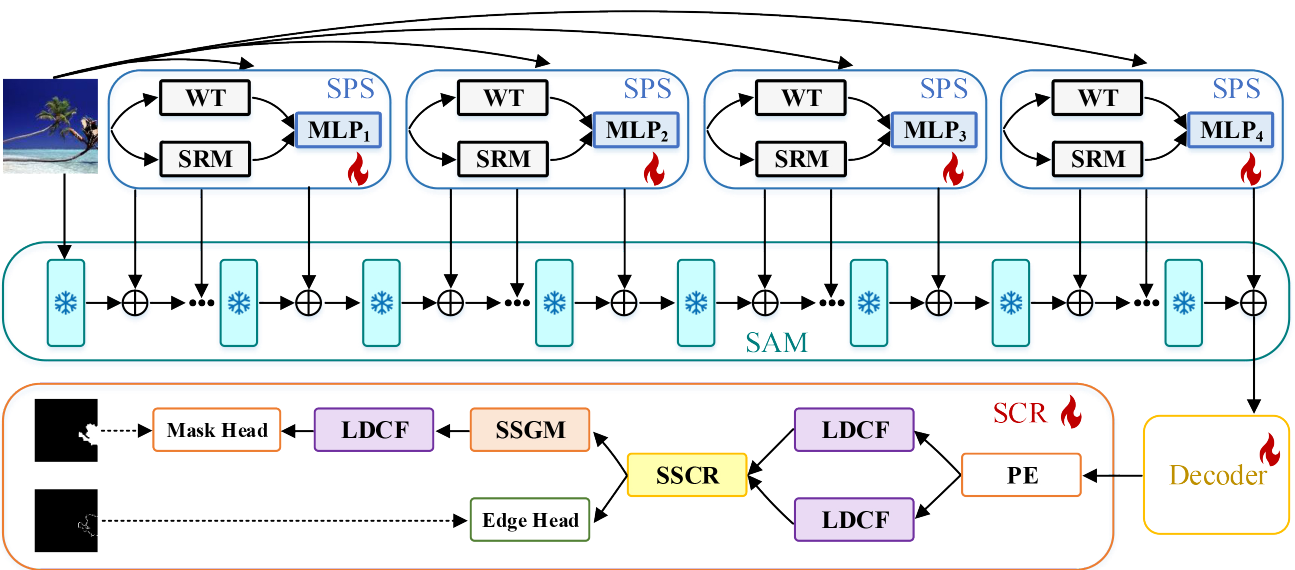}
		\caption{Framework of the proposed TRACE. ``WT'' denotes the wavelet transform. TRACE decomposes localization into three progressively coupled stages: semantic anchoring, semantic perturbation sensing, and semantic-constrained reasoning. Specifically, TRACE first grounds high-level interpretation to semantically meaningful objects and their spatial scope, then injects perturbation-sensitive frequency cues to capture subtle semantic edits, and finally refines the prediction through explicit semantic-scope reasoning.}
		\label{fig3}
	\end{figure*}
	
	\section{Approach}
	The overall framework is illustrated in Fig.~\ref{fig3}. Our method, termed TRACE, consists of three tightly coupled components: a semantic anchoring module, a semantic perturbation sensing module, and a semantic-constrained reasoning module. The three components play complementary roles: semantic anchoring determines where semantics concentrate, perturbation sensing identifies where subtle semantic changes may occur, and semantic-constrained reasoning determines which candidate regions remain coherent under joint content-scope reasoning.
	\subsection{Semantic Anchoring Module}
	The first challenge in SML is to identify the semantic support on which image interpretation mainly depends. Since semantic manipulations tend to occur in semantically decisive regions rather than arbitrary background areas, the model should first establish a reliable semantic anchor before searching for subtle perturbations. To this end, we adopt SAM3 \cite{carion2025sam,chen2025sam3} as the backbone of the semantic anchoring module. Compared with conventional forensic encoders that mainly emphasize low-level artifacts, SAM3 provides stronger object-level grouping priors and denser semantic region modeling, which are more suitable for localizing meaning-carrying image regions.
	
	We keep the SAM3 encoder frozen and train the adapter and decoder. This design preserves the strong visual prior of the foundation model while allowing task-specific adaptation to SML. We do not modify the decoder architecture; instead, the decoder directly outputs an initial coarse mask, which serves as a semantic scope initialization for subsequent refinement. Thus, the anchoring module does not aim to solve SML alone, but to provide a semantically grounded starting point for perturbation sensing and reasoning.
	
	\subsection{Semantic Perturbation Sensing Module}
	
	Semantic anchoring alone is insufficient for SML, because semantic edits often remain visually plausible and may not produce obvious anomalies in the spatial domain. In many cases, the manipulation is confined to a small attribute, state, or relation, while the surrounding context remains highly consistent\cite{yin2025adversarial,liu2025frequency}. As a result, purely semantic backbone features may identify the object but still under-represent weak perturbation evidence. To address this issue, we introduce a semantic perturbation sensing module that injects frequency-domain cues into the frozen backbone.
	
	Instead of relying only on global spectral representations, we combine wavelet decomposition and SRM filtering. For an input image $I=[I_R, I_G, I_B]$, we perform channel-wise wavelet analysis. For each channel $I_c$, where $c\in\{R,G,B\}$, we obtain
	\begin{equation}
		\mathcal{W}(I_c)=\{L_c,\; H_c^{h},\; H_c^{v},\; H_c^{d}\},
		\label{eq1}
	\end{equation}
	where $L_c$ is the low-frequency component and $H_c^{h}$, $H_c^{v}$, and $H_c^{d}$ denote the horizontal, vertical, and diagonal high-frequency subbands, respectively. Since subtle semantic manipulations are more likely to reside in local detail responses than in global smooth structures, we retain only the high-frequency terms:
	\begin{equation}
		F_c^{w}=\left[H_c^{h},\, H_c^{v},\, H_c^{d}\right].
		\label{eq2}
	\end{equation}
	The final wavelet representation is formed by aggregating all three channel-wise responses:
	\begin{equation}
		F^{w}=\left[F_R^{w},\, F_G^{w},\, F_B^{w}\right].
		\label{eq3}
	\end{equation}
	
	Different from previous methods, we deliberately perform the decomposition independently for each RGB channel rather than jointly over the three-channel input. This design is empirically more effective and also well motivated. First, subtle semantic edits often induce asymmetric chromatic perturbations, and early joint processing tends to wash out such channel-specific responses. Second, local high-frequency evidence in one channel may suppress weaker but meaningful responses in another when channels are mixed too early. Third, SML is sensitive to highly localized meaning-changing cues, for which preserving channel-level perturbation diversity is more beneficial than enforcing premature cross-channel coupling. In parallel, we apply SRM filtering to obtain residual-sensitive features:
	\begin{equation}
		F^{s}=\phi_{\mathrm{SRM}}(I).
		\label{eq4}
	\end{equation}
	The two streams are then fused into a unified perturbation representation
	\begin{equation}
		F^{p}=\left[F^{w},\, F^{s}\right].
		\label{eq5}
	\end{equation}
	Then, $F^{p}$ is projected into a prompt:
	\begin{equation}
		P^{i}=\mathrm{MLP}_{\mathrm{up}}\!\left(\mathrm{GELU}\!\left(\mathrm{MLP}_{\mathrm{down}}^{i}(F^{p})\right)\right),
		\label{eq6a}
	\end{equation}
	which is injected into the corresponding stage of the frozen SAM3 encoder. Here, $\mathrm{MLP}_{\mathrm{down}}^{i}$ is stage-specific, while $\mathrm{MLP}_{\mathrm{up}}$ is used to align the prompt dimension with the transformer feature space. In this way, the backbone becomes sensitive not only to semantic support, but also to subtle local perturbations that may trigger semantic shifts. The resulting adapted features are decoded by the trainable SAM3 mask decoder to obtain an initial coarse mask $M_{0}$, which is further refined by the semantic-constrained reasoning module.
	
	\subsection{Semantic-Constrained Reasoning Module}
	
	The decoder output $M_{0}$ provides a coarse semantic initialization, but SML requires more than region-level grounding or boundary refinement. A semantically manipulated region is often small, visually consistent, and only meaningful when interpreted together with its surrounding semantic scope. Therefore, the model should not merely sharpen local responses, but explicitly determine whether a candidate region is supported by both \emph{what} semantic content is being altered and \emph{where} the valid semantic scope of this alteration lies. For this reason, we cast the refinement stage as semantic-constrained reasoning rather than conventional mask enhancement.
	
	To achieve this, we introduce a semantic-constrained reasoning module based on Mamba. The key idea is to cast refinement as a process of \emph{candidate verification} rather than direct feature enhancement. In SML, a local response should not be accepted merely because it is strong; it should be preserved only if it remains semantically valid when examined together with both the manipulated content and its surrounding semantic scope. Therefore, our module explicitly treats candidate regions as semantic hypotheses and verifies them under coupled content-scope evidence. The module performs reasoning in three steps. First, it decomposes the coarse prediction into a content branch and a scope branch, such that each candidate region is represented from two complementary perspectives: \emph{what} semantic content may have been altered and \emph{where} the valid semantic scope of this alteration lies. Second, it interleaves the two branches and propagates them jointly along multiple scan paths, so that the same candidate hypothesis is repeatedly examined under different contextual traversals. Third, it consolidates the multi-directional responses and feeds the verified scope cues back to the content branch for final correction. In this way, the module does not simply aggregate local cues, but progressively verifies, rejects, and consolidates semantic hypotheses under joint content-scope constraints. The module contains three parts: Local Decisive Cue Fusion (LDCF), Semantic-Scope Coherence Reasoning (SSCR), and Semantic Scope-Guided Modulation (SSGM).
	
	\paragraph{Local Decisive Cue Fusion}
	Starting from the initial decoder prediction $M_{0}$, we first transform it into a compact latent representation through patch embedding. On top of this embedded feature, we apply a lightweight local fusion block, termed \emph{Local Decisive Cue Fusion} (LDCF), before entering the reasoning stage. Concretely, LDCF is implemented as a shallow convolutional block consisting of a convolution layer, followed by batch normalization and a gated linear unit, and then another convolution layer with a GELU activation. Rather than serving as a heavy feature transformation module, LDCF is designed to perform compact local context aggregation and cue refinement with minimal overhead.
	
	The role of LDCF is twofold. First, it consolidates short-range neighborhood evidence around the coarse prediction $M_{0}$, which is important because semantic manipulations are often highly localized and their decisive cues may only emerge in small spatial neighborhoods. Second, it produces two complementary feature streams for the subsequent reasoning module: a mask stream that emphasizes semantic content and an edge stream that captures semantic scope. We denote these two features by $F_{m}$ and $F_{e}$, respectively. They are then serialized to form the input sequences of SSCR. This is why we term it \emph{Local Decisive Cue Fusion}: it does not perform reasoning by itself, but extracts and stabilizes the local decisive cues on which the subsequent semantic-constrained reasoning operates.
	
	\begin{figure*}[t]
		\centering
		\includegraphics[width=\textwidth]{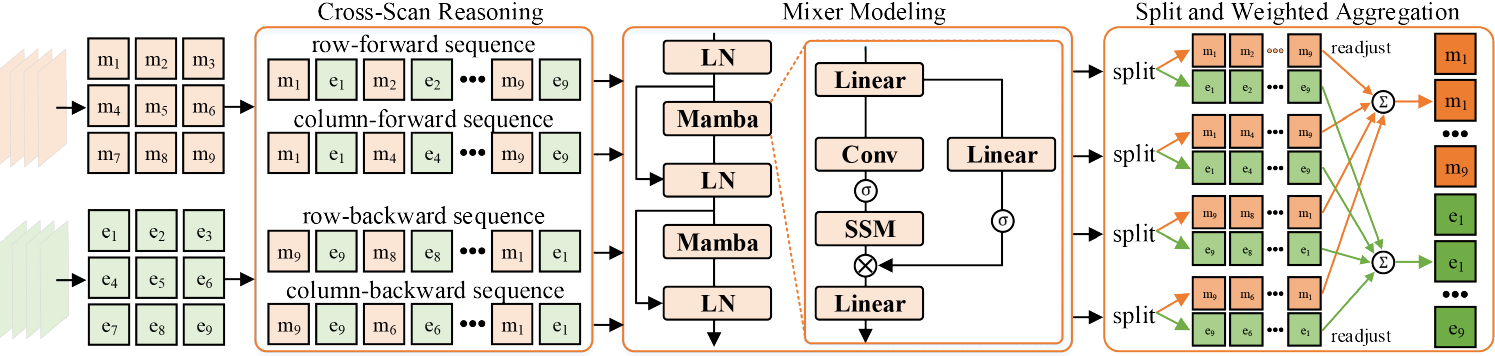}
		\caption{Framework of the semantic-scope coherence reasoning (SSCR). Starting from the mask and edge features, SSCR first constructs four interleaved semantic-scope sequences under row-forward, column-forward, row-backward, and column-backward scan orders (cross-scan reasoning). These sequences are then processed by a shared MixerModel composed of stacked Mamba (mixer modeling). Finally, each directional output is split by position back into mask and edge subsequences, and the four directional responses are aggregated separately with learnable channel-wise weights to obtain the refined mask and edge features (split and weighted aggregation).}
		\label{fig4}
	\end{figure*}
	
	\paragraph{Semantic-Scope Coherence Reasoning}
	Built upon the mask and edge features $F_{m}$ and $F_{e}$ extracted from the coarse prediction $M_{0}$ by LDCF, SSCR serves as the core reasoning component of our framework. The key idea is that a candidate region should be judged by the joint consistency of semantic content and semantic scope, rather than by either cue alone. To this end, SSCR explicitly couples the two branches during sequence propagation, so that each update is conditioned on both what may have been altered and where its valid semantic scope lies. Moreover, by propagating the coupled representation along multiple scan paths, SSCR repeatedly examines the same candidate under different contextual traversals, suppressing responses that are only locally strong but semantically inconsistent while preserving those that remain coherent across content-scope interactions.
	
	Specifically, the mask feature $F_{m}$ and the edge feature $F_{e}$ are first serialized into two token sequences,
	\begin{equation}
		\begin{aligned}
			S_{m} &= \left[m_{1},\, m_{2},\, \dots,\, m_{N}\right], \\
			S_{e} &= \left[e_{1},\, e_{2},\, \dots,\, e_{N}\right],
		\end{aligned}
		\label{eq9}
	\end{equation}
	where $S_{m}$ represents the sequence form of the semantic content feature derived from the coarse prediction $M_{0}$, while $S_{e}$ represents the corresponding semantic-scope feature. As illustrated in Fig.~\ref{fig4}, SSCR directly constructs four interleaved semantic-scope sequences under four scan paths, namely row-forward, column-forward, row-backward, and column-backward:
	\begin{equation}
		\left\{S^{(r+)},\, S^{(c+)},\, S^{(r-)},\, S^{(c-)}\right\}.
		\label{eq10}
	\end{equation}
	For each direction, the corresponding mixed sequence is formed by alternating mask and edge tokens according to the scan order:
	\begin{equation}
		S^{(d)}=\left[m_{\pi_d(1)},\, e_{\pi_d(1)},\, m_{\pi_d(2)},\, e_{\pi_d(2)},\, \dots,\, m_{\pi_d(N)},\, e_{\pi_d(N)}\right],
		\label{eq11}
	\end{equation}
	where $\pi_d(\cdot)$ denotes the index permutation induced by direction $d\in\{r+,c+,r-,c-\}$. In this way, the same candidate region is examined together with its scope cue under different contextual traversals.
	
	These four interleaved sequences are then fed into a shared mixer modeling, as shown in the middle part of Fig.~\ref{fig4}. The mixer modeling is composed of stacked Mamba-based blocks and performs joint semantic-scope propagation along each scan path. For each direction $d$, the output sequence is $\widehat{S}^{(d)}=\mathcal{M}\!\left(S^{(d)}\right)$. During this process, semantic content and semantic scope are jointly propagated along each scan path, so that candidate responses are repeatedly examined under different contextual traversals. Since each update is jointly conditioned on content and scope, the propagation process does not merely enhance local responses, but progressively enforces semantic coherence along each scan path.
	
	After mixer modeling, each directional output remains in the interleaved form. Therefore, as shown in the right part of Fig.~\ref{fig4}, we first split each sequence by position back into a mask subsequence and an edge subsequence:
	\begin{equation}
		\widehat{S}^{(d)} \rightarrow \left(\widehat{S}_{m}^{(d)},\, \widehat{S}_{e}^{(d)}\right).
		\label{eq14}
	\end{equation}
	The split features are then aggregated separately through learnable directional weights. For each channel $c$, we perform weighted aggregation over the four directions:
	\begin{equation}
		\begin{aligned}
			\overline{S}_{m}(:,c) &= \sum_{d} w_{d,c}^{m}\,\widehat{S}_{m}^{(d)}(:,c), \\
			\overline{S}_{e}(:,c) &= \sum_{d} w_{d,c}^{e}\,\widehat{S}_{e}^{(d)}(:,c),
		\end{aligned}
		\label{eq15}
	\end{equation}
	where $w_{d,c}^{m}$ and $w_{d,c}^{e}$ are learnable parameters. The aggregation is channel-wise, so each channel learns its own directional preference over the four scan paths. The aggregated sequences are finally reshaped back into 2D features, denoted by $M_{f}$ and $E_{f}$.
	
	\paragraph{Semantic Scope-Guided Modulation}
	After SSCR, we obtain an aggregated mask feature $M_{f}$ and an aggregated edge feature $E_{f}$. Although the edge branch provides informative scope cues, not all of them are equally useful for final localization. In SML, the valid contribution of scope evidence should depend on the current semantic hypothesis, rather than being injected uniformly. This motivates SSGM: instead of directly fusing the edge feature into the mask branch, we let the mask feature generate a gate to selectively modulate the scope feature. In this way, the model can retain scope cues that are semantically consistent with the current candidate region, while suppressing irrelevant or noisy structural responses.
	
	Concretely, the mask feature generates a gating map, which is then used to modulate the edge feature before fusion:
	\begin{equation}
		\begin{aligned}
			G &= \sigma\!\left(\mathrm{Linear}\!\left(\mathrm{LN}(M_{f})\right)\right), \\
			M_{c} &= M_{f}+G\odot E_{f},
		\end{aligned}
		\label{eq16}
	\end{equation}
	where $G$ denotes the learned modulation gate, and $\sigma(\cdot)$ is the sigmoid activation. This design is semantically motivated: the mask branch encodes the current estimate of manipulated content, and thus is better suited to decide which scope cues should be preserved for refinement. Therefore, SSGM acts as a semantic relevance filter, allowing scope information to correct the content branch in a selective rather than unconditional manner.
	
	After modulation, we further apply a second LDCF block for local consolidation before final decoding. This step complements the global reasoning of SSCR by re-introducing compact neighborhood fusion, allowing the refined representation to better recover local structural consistency around the manipulated region. A mask head is then applied to produce the final prediction $M$, while the edge feature is fed into an edge head to produce the auxiliary edge prediction $E$.
	
	\subsection{Training Objective}
	We supervise the final mask prediction using the combination of binary cross-entropy and IoU loss, while the auxiliary edge branch is supervised by binary cross-entropy only. Specifically, the mask loss is defined as
	\begin{equation}
		\mathcal{L}_{\mathrm{mask}}
		=
		\mathcal{L}_{\mathrm{BCE}}^{m}
		+
		\mathcal{L}_{\mathrm{IoU}}^{m},
		\label{eq19}
	\end{equation}
	where
	\begin{equation}
		\mathcal{L}_{\mathrm{BCE}}^{m}
		=
		-\sum_{i}
		\left(
		T_{i}\log M_{i}
		+
		(1-T_{i})\log(1-M_{i})
		\right),
		\label{eq20}
	\end{equation}
	and the IoU loss is written as
	\begin{equation}
		\mathcal{L}_{\mathrm{IoU}}^{m}
		=
		1-
		\frac{\sum_{i} T_{i} M_{i}}
		{\sum_{i} T_{i}+\sum_{i} M_{i}-\sum_{i} T_{i} M_{i}},
		\label{eq21}
	\end{equation}
	where $T_{i}$ denotes the ground-truth mask label at pixel $i$, and $M_{i}$ denotes the corresponding predicted mask probability.
	
	For the edge branch, we use binary cross-entropy loss:
	\begin{equation}
		\mathcal{L}_{\mathrm{edge}}
		=
		-\sum_{i}
		\left(
		Q_{i}\log E_{i}
		+
		(1-Q_{i})\log(1-E_{i})
		\right),
		\label{eq22}
	\end{equation}
	where $Q_{i}$ and $E_{i}$ denote the ground-truth and predicted edge labels at pixel $i$, respectively. The overall training objective is
	\begin{equation}
		\mathcal{L}
		=
		\mathcal{L}_{\mathrm{mask}}
		+
		\mathcal{L}_{\mathrm{edge}}.
		\label{eq23}
	\end{equation}
	
	\begin{table*}[t]
		\centering
		\scriptsize
		\setlength{\tabcolsep}{3.5pt}
		\caption{Quantitative comparison on the full test set and on the dataset constructed using the \emph{Semantically Decisive Region Identification and Semantic Tampering (SDRI-ST)} pipeline. All values are rounded to three decimals. The best results are shown in \textbf{bold}, and the second-best results are underlined. RI denotes the relative improvement of our method over the second-best method, computed as $(\mathrm{Ours}-\mathrm{Second})/\mathrm{Second}$ for all metrics except MAE, for which we use $(\mathrm{Second}-\mathrm{Ours})/\mathrm{Second}$ since lower is better.}
		\label{tab:quantitative_comparison_combined}
		\begin{tabular}{l|ccccccc|ccccccc}
			\toprule
			\multirow{2}{*}{Method} & \multicolumn{7}{c|}{Full Test Set} & \multicolumn{7}{c}{SDRI+ST Dataset} \\
			& IoU & Dice & Precision & Recall & Accuracy & MAE & AUC & IoU & Dice & Precision & Recall & Accuracy & MAE & AUC \\
			\midrule
			MVSS-Net\cite{dong2022mvss}            & 0.601 & 0.673 & 0.855 & 0.648 & 0.934 & 0.066 & 0.926 & 0.622 & 0.702 & 0.733 & 0.743 & 0.962 & 0.038 & 0.913 \\
			CFL-Net\cite{niloy2023cfl}             & 0.459 & 0.578 & 0.638 & 0.616 & 0.898 & 0.102 & 0.895 & 0.467 & 0.571 & 0.548 & 0.689 & 0.938 & 0.062 & 0.883 \\
			IML-ViT\cite{ma2023iml}                & 0.672 & 0.737 & \underline{0.935} & 0.702 & 0.944 & 0.056 & \underline{0.974} & 0.680 & 0.761 & 0.781 & 0.797 & 0.969 & 0.031 & 0.953 \\
			EITLNet\cite{guo2024effective}         & 0.635 & 0.694 & 0.913 & 0.678 & 0.936 & 0.064 & 0.955 & \underline{0.742} & 0.802 & \underline{0.840} & 0.812 & \underline{0.978} & \underline{0.022} & 0.959 \\
			MMFusion-IML\cite{triaridis2023mmfusion} & 0.693 & \underline{0.779} & 0.855 & \underline{0.792} & 0.946 & 0.054 & 0.966 & 0.716 & 0.799 & 0.787 & \underline{0.869} & 0.968 & 0.032 & \underline{0.977} \\
			HDFNet\cite{han2024hdf}                & 0.462 & 0.582 & 0.661 & 0.634 & 0.891 & 0.109 & 0.878 & 0.353 & 0.467 & 0.419 & 0.655 & 0.888 & 0.112 & 0.854 \\
			SAM2-UNET\cite{xiong2026sam2}          & \underline{0.702} & 0.769 & 0.917 & 0.754 & \underline{0.946} & \underline{0.054} & 0.942 & \underline{0.742} & \underline{0.812} & 0.816 & 0.863 & 0.975 & 0.025 & 0.969 \\
			Trace                                  & \textbf{0.789} & \textbf{0.846} & \textbf{0.941} & \textbf{0.828} & \textbf{0.960} & \textbf{0.040} & \textbf{0.976} & \textbf{0.847} & \textbf{0.896} & \textbf{0.897} & \textbf{0.920} & \textbf{0.984} & \textbf{0.016} & \textbf{0.985} \\
			\midrule
			RI                                     & 12.391\% & 8.555\% & 0.615\% & 4.594\% & 1.472\% & 25.822\% & 0.244\% & 14.129\% & 10.375\% & 6.773\% & 5.860\% & 0.613\% & 25.567\% & 0.861\% \\
			\bottomrule
		\end{tabular}
	\end{table*}
	\section{Experiments}
	This section presents a systematic experimental evaluation of the proposed network. It covers the datasets, implementation details, quantitative and qualitative comparisons with image manipulation localization methods, ablation studies, and corresponding visualization analyses, aiming to comprehensively validate the effectiveness of the proposed approach.
	\subsection{Experiment Setup}
	\paragraph{Implementation Details}
	This work is implemented in the PyTorch framework, and the proposed model Trace is trained on a single Ascend 910B NPU. All images and masks are uniformly resized to 
	$1008\times 1008$. Both the input images and the ground-truth masks are normalized before being fed into the network. During training, the model further leverages pre-computed edge information to facilitate learning, where the edge maps are extracted using the Sobel operator. We use AdamW as the optimizer, with an initial learning rate of $2\times 10^{-4}$, and a minimum learning rate of $1\times 10^{-7}$. The batch size is set to 4, and the model is trained for 20 epochs. The parameters of the encoder backbone are kept frozen throughout training.
	
	\paragraph{Evaluation Metrics}
	We evaluate the proposed method using seven metrics, including IoU, Dice, Precision, Recall, Accuracy, AUC, and MAE. Let $T_i \in \{0,1\}$ denote the ground-truth label of pixel $i$, $M_i \in [0,1]$ denote the predicted probability, and $\widetilde{M}_i \in \{0,1\}$ denote the binarized prediction.
	\begin{equation}
		\begin{aligned}
			\mathrm{IoU} &=
			\frac{\sum_i T_i \widetilde{M}_i}
			{\sum_i T_i+\sum_i \widetilde{M}_i-\sum_i T_i \widetilde{M}_i}, \\
			\mathrm{Dice} &=
			\frac{2\sum_i T_i \widetilde{M}_i}
			{\sum_i T_i+\sum_i \widetilde{M}_i}, \\
			\mathrm{Precision} &=
			\frac{\sum_i T_i \widetilde{M}_i}
			{\sum_i \widetilde{M}_i}, \\
			\mathrm{Recall} &=
			\frac{\sum_i T_i \widetilde{M}_i}
			{\sum_i T_i}, \\
			\mathrm{Accuracy} &=
			\frac{1}{N}\sum_i \mathbb{I}(T_i=\widetilde{M}_i), \\
			\mathrm{AUC} &=
			\int_0^1 \mathrm{TPR}(t)\, d\mathrm{FPR}(t), \\
			\mathrm{MAE} &=
			\frac{1}{N}\sum_i |M_i-T_i|,
		\end{aligned}
		\label{eq24}
	\end{equation}
	where $N$ is the total number of pixels, $\mathbb{I}(\cdot)$ is the indicator function, and $\mathrm{TPR}(t)$ and $\mathrm{FPR}(t)$ denote the true positive rate and false positive rate at threshold $t$, respectively.

	\begin{figure*}[t]
		\centering
		\includegraphics[width=\textwidth]{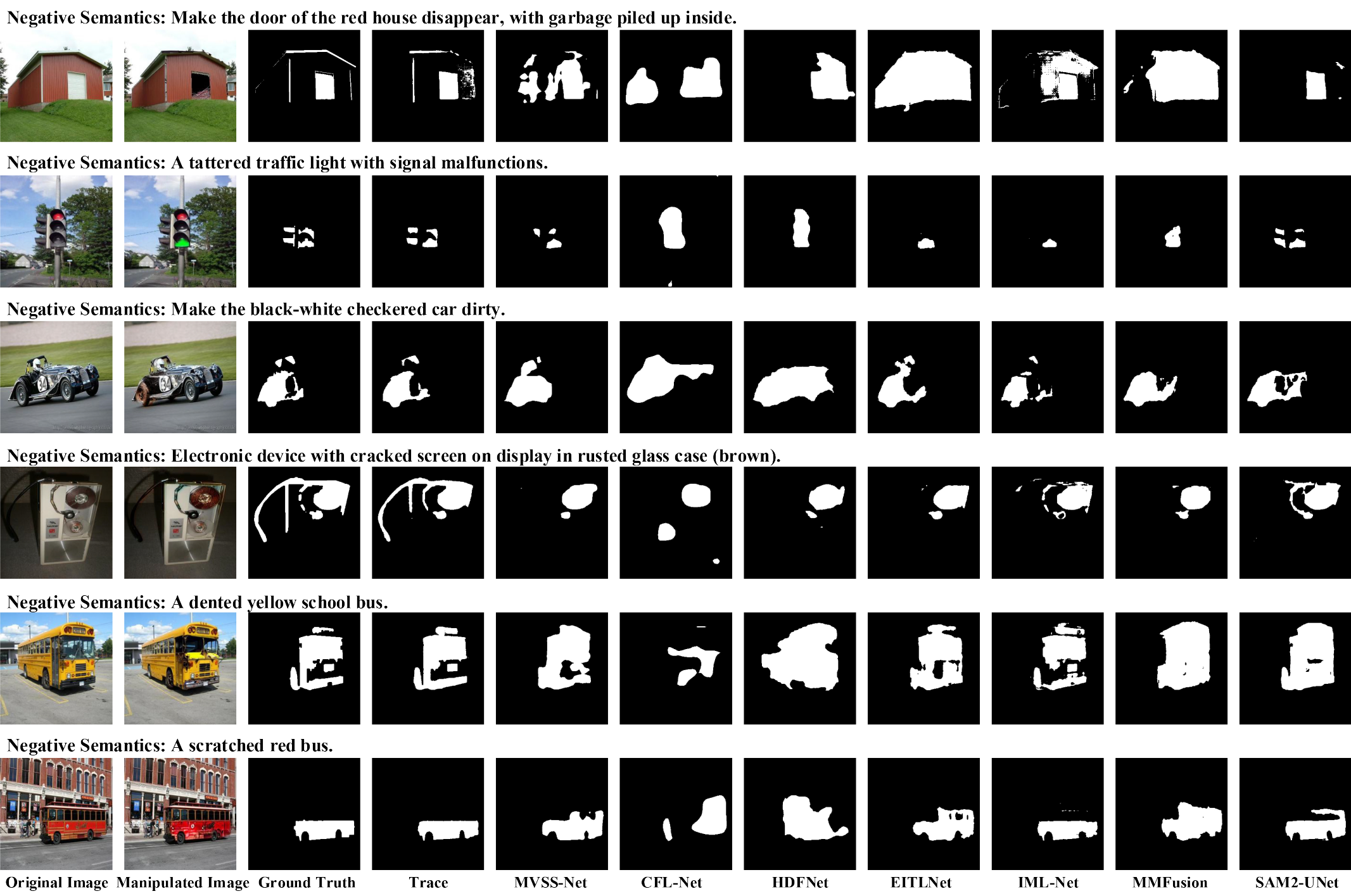}
		\caption{Qualitative results of our Trace compared with other SoTAs on our proposed dataset. In addition, We provide negative semantics generated by LLMs to help clarify how cognitive changes occur in images.}
		\label{fig5}
	\end{figure*}
	
	\subsection{Comparison with SOTAs}
	\paragraph{Compared Methods}
	We evaluate and compare our Trace with seven SoTA methods that release codes, including MVSS-Net\cite{dong2022mvss}, CFL-Net\cite{niloy2023cfl}, IML-ViT\cite{ma2023iml}, EITLNet\cite{guo2024effective}, MMFusion\cite{triaridis2023mmfusion}, HDFNet\cite{han2024hdf}, SAM2-UNET\cite{xiong2026sam2}. For a fair comparison, we retrain these methods using our datasets under the same settings.
	
	\paragraph{Quantitative Results}
	Tab.~\ref{tab:quantitative_comparison_combined} shows that Trace achieves the best results on all seven metrics on both datasets. More importantly, its advantage is not merely a marginal improvement over the previous best method. On the first dataset, the gap between the second-best and third-best methods is relatively small for most key metrics, $\textit{e.g.}$, only $0.009$ in IoU and $0.010$ in Dice, whereas Trace improves over the second-best method by $0.087$ IoU and $0.067$ Dice. A similar trend can be observed for Recall and MAE, where the previous methods differ by only a limited margin, but Trace still brings a clear additional gain. This indicates that the advantage of Trace does not come from a favorable training setup, but from its stronger ability to distinguish semantically valid manipulation regions from visually similar but irrelevant responses.
	
	The improvement becomes even more pronounced on the dataset constructed by the \emph{Semantically Decisive Region Identification and Semantic Tampering} pipeline. Although this test set contains fewer images, which makes its overall absolute metric values slightly higher due to lower distribution diversity, it is actually more challenging in terms of semantic localization because the manipulations are intentionally concentrated in semantically decisive regions and remain highly consistent with the surrounding content. As a result, traditional IML-oriented methods, which mainly depend on low-level inconsistencies, become less reliable in this setting. This is also reflected in the comparison margins: for example, the gap between the second-best and third-best methods is almost negligible in IoU ($0.742$ vs.\ $0.742$ after rounding), while Trace further improves IoU to $0.847$; similarly, the previous margin in Dice is only about $0.010$, whereas Trace improves over the second-best method by $0.084$. The same pattern also appears in Precision, Recall, Accuracy, MAE, and AUC. These larger gains are consistent with our design motivation. Semantic anchoring helps the model first focus on meaning-carrying regions, semantic perturbation sensing enhances sensitivity to subtle yet meaningful edits, and semantic-constrained reasoning further verifies candidate regions under joint content-scope constraints. Therefore, when the benchmark emphasizes semantically decisive and visually consistent manipulations, the advantage of Trace becomes more evident.
	
	\paragraph{Qualitative Results}
	The qualitative comparisons further confirm the advantage of Trace. Compared with previous methods, our predictions are generally more complete, more compact, and better aligned with the true manipulated regions, especially when the edited area is small or lies inside a semantically important object. Conventional methods often exhibit three typical failure modes: missing decisive local regions, over-activating irrelevant context, and producing fragmented or semantically incoherent masks. These failures are expected, since methods designed for traditional IML tend to focus on visible low-level artifacts rather than semantic impact. By contrast, Trace produces cleaner and more semantically meaningful localization results. This improvement can be attributed to two aspects. First, the semantic anchoring and perturbation sensing modules make the model focus on regions that are both semantically important and potentially manipulated. Second, the semantic-constrained reasoning module jointly propagates semantic content and semantic scope, allowing the model to retain only those candidate responses that remain coherent under multiple contextual traversals. As a result, Trace yields sharper boundaries, fewer false positives, and stronger robustness on challenging semantic manipulation cases.
	\begin{figure*}[t]
		\centering
		\includegraphics[width=\textwidth]{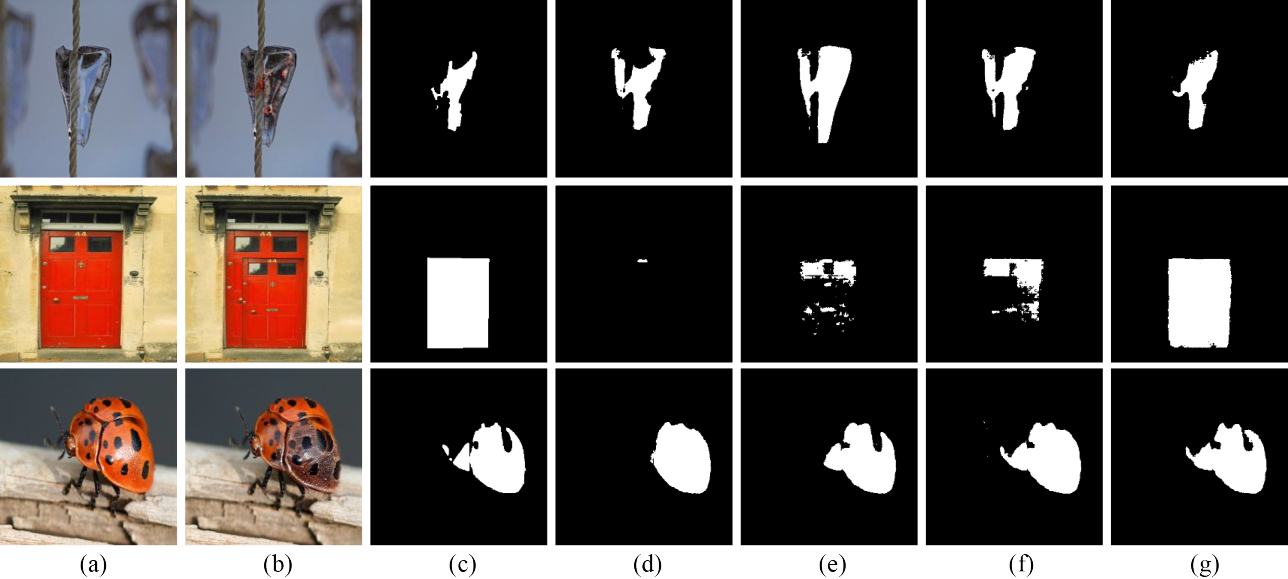}
		\caption{Qualitative results of each proposed module. (a) Original image; (b) Manipulated image; (c) Ground truth; (d) Baseline; (e) Baseline + SPS; (f) Baseline + Mask; (g) Baseline + SCR.}
		\label{fig6}
	\end{figure*}
	\begin{table*}[t]
		\centering
		\small
		\setlength{\tabcolsep}{5pt}
		\caption{Ablation study of the proposed framework. We progressively evaluate the contribution on semantic manipulation localization.}
		\label{tab:ablation}
		\begin{tabular}{c|c|cc|cc|ccccccc}
			\toprule
			\multirow{2}{*}{ID} & \multirow{2}{*}{SAM} & \multicolumn{2}{c|}{SPS} & \multicolumn{2}{c|}{SCR} & \multirow{2}{*}{IoU} & \multirow{2}{*}{Dice} & \multirow{2}{*}{Precision} & \multirow{2}{*}{Recall} & \multirow{2}{*}{Accuracy} & \multirow{2}{*}{MAE} & \multirow{2}{*}{AUC} \\
			&  & Wavelet & SRM & Mask & Edge &  &  &  &  &  &  &  \\
			\midrule
			1 &  &  &  &  &  & 0.293 & 0.420 & 0.312 & 0.879 & 0.631 & 0.369 & 0.817 \\
			2 &\checkmark  &  &  &  &  & 0.749 & 0.803 & 0.932 & 0.783 & 0.951 & 0.049 & 0.964 \\
			3 &\checkmark  & \checkmark &  &  &  & 0.750 & 0.804 & 0.938 & 0.783 & 0.951 & 0.049 & 0.966 \\
			4 &\checkmark  &  & \checkmark &  &  & 0.764 & 0.820 & 0.929 & 0.801 & 0.953 & 0.047 & 0.972 \\
			5 &\checkmark  & \checkmark & \checkmark &  &  & 0.770 & 0.826 & 0.923 & 0.809 & 0.955 & 0.045 & 0.974 \\
			6 &\checkmark  & \checkmark & \checkmark & \checkmark &  & 0.767 & 0.828 & 0.942 & 0.796 & 0.955 & 0.045 & 0.971 \\
			7 &\checkmark  & \checkmark & \checkmark &  & \checkmark & 0.764 & 0.822 & 0.932 & 0.804 & 0.954 & 0.046 & 0.971 \\
			8 &\checkmark  & \checkmark & \checkmark & \checkmark & \checkmark & \textbf{0.789} & \textbf{0.846} & \textbf{0.941} & \textbf{0.828} & \textbf{0.960} & \textbf{0.040} & \textbf{0.976} \\
			\bottomrule
		\end{tabular}
	\end{table*}
	\subsection{Ablation Study}
	In this section, we conduct ablation studies of SAM, SPS and SCR. Tab.~\ref{tab:ablation} and Fig.~\ref{fig6} report and show the results. Note, in Tab.~\ref{tab:ablation} and Fig.~\ref{fig6}, the individual \emph{Mask} and \emph{Edge} entries indicate that the Mamba operation is performed using only the mask feature or only the edge feature, respectively, without the \emph{Cross-Scan Reasoning} in SSCR or the SSGM operation. Only when both \emph{Mask} and \emph{Edge} are included does the full SSCR and SSGM design take effect.
	
	\paragraph{Effectiveness of SAM}
	We first examine the contribution of the semantic anchoring module. Compared with a basic U-Net style baseline, introducing SAM leads to a clear and substantial performance improvement. Unlike conventional segmentation backbones that mainly rely on local appearance aggregation, SAM provides stronger object-level grouping priors and denser semantic region modeling, which are much better aligned with the nature of SML, where manipulations are typically embedded in semantically meaningful regions rather than arbitrary background areas. Therefore, SAM does not simply improve feature quality in a generic sense; more importantly, it offers a semantically grounded initialization that allows the subsequent modules to refine predictions from a much more reliable starting point.
	
	\begin{figure}[t]
		\centering
		\includegraphics[width=0.48\textwidth]{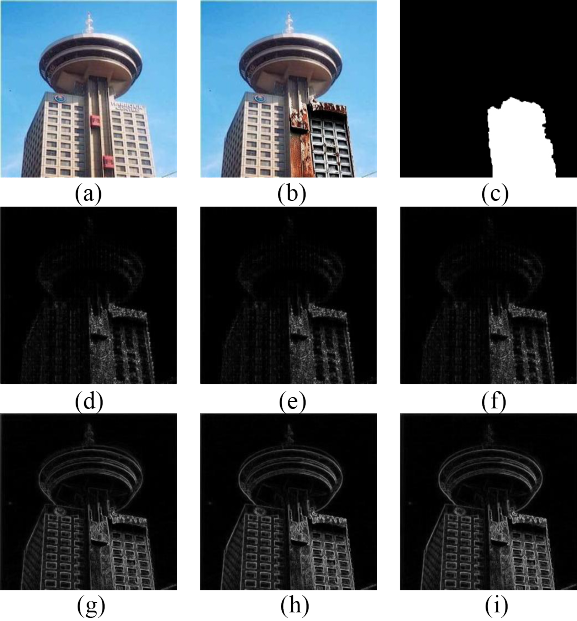}
		\caption{Visualization results of SPS. (a) Original image; (b) Manipulated image; (c) Ground truth; (d) SRM result of `R' channel; (e) SRM result of `G' channel; (f) SRM result of `B' channel; (g) Wavelet result of `R' channel; (h) Wavelet result of `G' channel; (i) Wavelet result of `B' channel.}
		\label{fig7}
	\end{figure}
	
	\paragraph{Effectiveness of SPS}
	We then evaluate the effectiveness of the semantic perturbation sensing module. As shown in Tab.~\ref{tab:ablation}, adding either wavelet cues or SRM cues on top of SAM brings further improvement, and combining both yields a more consistent gain than using either one alone. This verifies that semantic anchoring by itself is still insufficient for SML. Although SAM can provide strong semantic support, semantically manipulated regions are often highly localized and visually consistent with the surrounding context, so their distinguishing evidence may remain weak in purely semantic backbone features. SPS addresses this issue by injecting perturbation-sensitive frequency cues into the frozen backbone.
	
	The improvement pattern is also meaningful. Wavelet decomposition and SRM filtering capture different aspects of subtle perturbations. The wavelet branch focuses on localized high-frequency responses and preserves channel-wise perturbation diversity, while the SRM branch emphasizes residual-sensitive forensic traces. Their combination is therefore complementary rather than redundant. This complementarity can be observed in Fig.~\ref{fig7}. The SRM responses highlight suspicious residual structures, whereas the wavelet responses reveal local high-frequency variations across different channels. They provide useful evidence for prompting the backbone toward subtle semantic edits that would otherwise be suppressed by strong visual consistency. As a result, SPS improves not only the sensitivity to manipulated regions, but also the reliability of subsequent refinement.

	\paragraph{Effectiveness of SCR}
	We analyze the contribution of the SCR module, which is the core refinement component of TRACE. Tab.~\ref{tab:ablation} shows that simply introducing a single-branch Mamba refinement with only the mask feature or only the edge feature brings limited or unstable gains, whereas enabling the full SCR design with both branches leads to the best overall result. This demonstrates that the advantage of SCR does not come merely from adding a stronger sequence model. Instead, its key value lies in explicitly modeling the interaction between semantic content and semantic scope. In SML, a candidate region should not be preserved only because its local response is strong. It should be retained only when the manipulated content remains coherent with its valid semantic scope. This is exactly the role of SCR.
	
	\begin{figure}[t]
		\centering
		\includegraphics[width=0.48\textwidth]{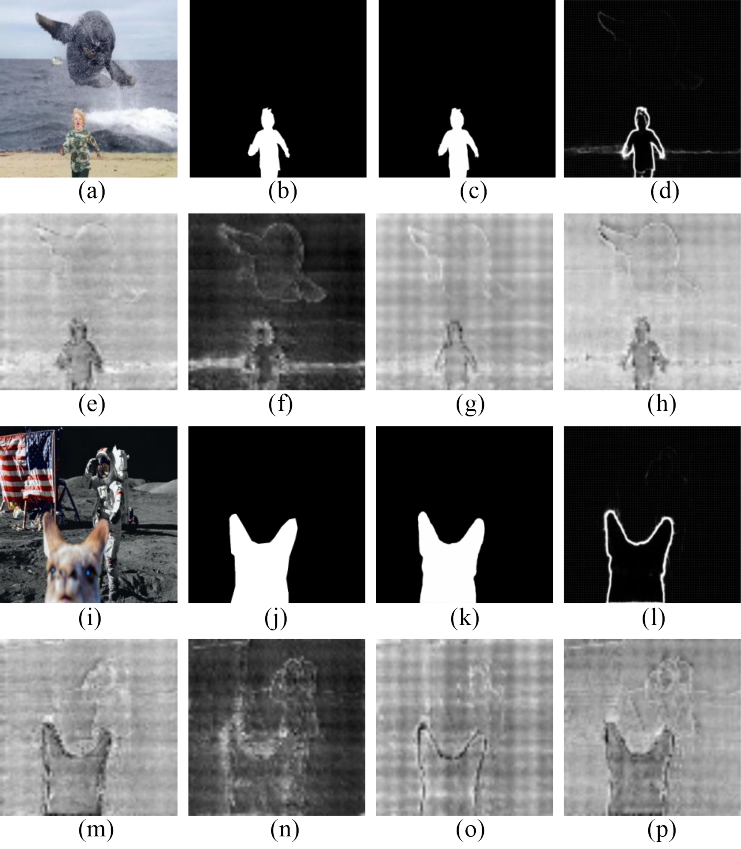}
		\caption{Visualization results of SSCR. (a) (i) Manipulated image; (b) (j) Ground truth; (c) (k) Predicted result; (d) (l) Predicted edge; (e) (m) Visualization feature of row-forward sequence; (f) (n) Visualization feature of row-backward sequence; (g) (o) Visualization feature of column-forward sequence; (h) (p) Visualization feature of column-backward sequence.}
		\label{fig8}
	\end{figure}
	
	\begin{table*}[t]
		\centering
		\small
		\setlength{\tabcolsep}{6pt}
		\renewcommand{\arraystretch}{1.15}
		\caption{Robustness comparison in terms of IoU under different post-processing perturbations. The best results are shown in bold.}
		\label{tab:robustness_iou}
		\begin{tabular}{llcccccccc}
			\toprule
			\multirow{2}{*}{\begin{tabular}[c]{@{}l@{}}Post-processing\\ method\end{tabular}}
			& \multirow{2}{*}{\begin{tabular}[c]{@{}l@{}}Parameter\\ value\end{tabular}}
			& \multicolumn{8}{c}{Methods} \\
			\cmidrule(lr){3-10}
			& & Ours & SAM2-UNET & CFL-Net & EITLNet & HDFNet & IML-ViT & MMFunsion & MVSSnet \\
			\midrule
			\multirow{3}{*}{Gaussian blur}
			& kernel=3  & \textbf{0.790} & 0.532 & 0.225 & 0.319 & 0.433 & 0.288 & 0.510 & 0.132 \\
			& kernel=9  & \textbf{0.745} & 0.315 & 0.251 & 0.187 & 0.432 & 0.215 & 0.436 & 0.081 \\
			& kernel=15 & \textbf{0.666} & 0.285 & 0.259 & 0.160 & 0.423 & 0.295 & 0.408 & 0.100 \\
			\midrule
			\multirow{2}{*}{JPEG compression}
			& quality=50 & \textbf{0.717} & 0.657 & 0.417 & 0.576 & 0.457 & 0.569 & 0.597 & 0.462 \\
			& quality=75 & \textbf{0.735} & 0.663 & 0.448 & 0.612 & 0.461 & 0.630 & 0.630 & 0.520 \\
			\midrule
			\multirow{3}{*}{Gaussian noise}
			& sigma=3  & \textbf{0.767} & 0.667 & 0.450 & 0.621 & 0.460 & 0.642 & 0.662 & 0.558 \\
			& sigma=9  & \textbf{0.738} & 0.648 & 0.428 & 0.586 & 0.460 & 0.629 & 0.624 & 0.509 \\
			& sigma=15 & \textbf{0.719} & 0.628 & 0.386 & 0.512 & 0.455 & 0.581 & 0.530 & 0.426 \\
			\bottomrule
		\end{tabular}
	\end{table*}
	
	\begin{figure}[t]
		\centering
		\includegraphics[width=0.48\textwidth]{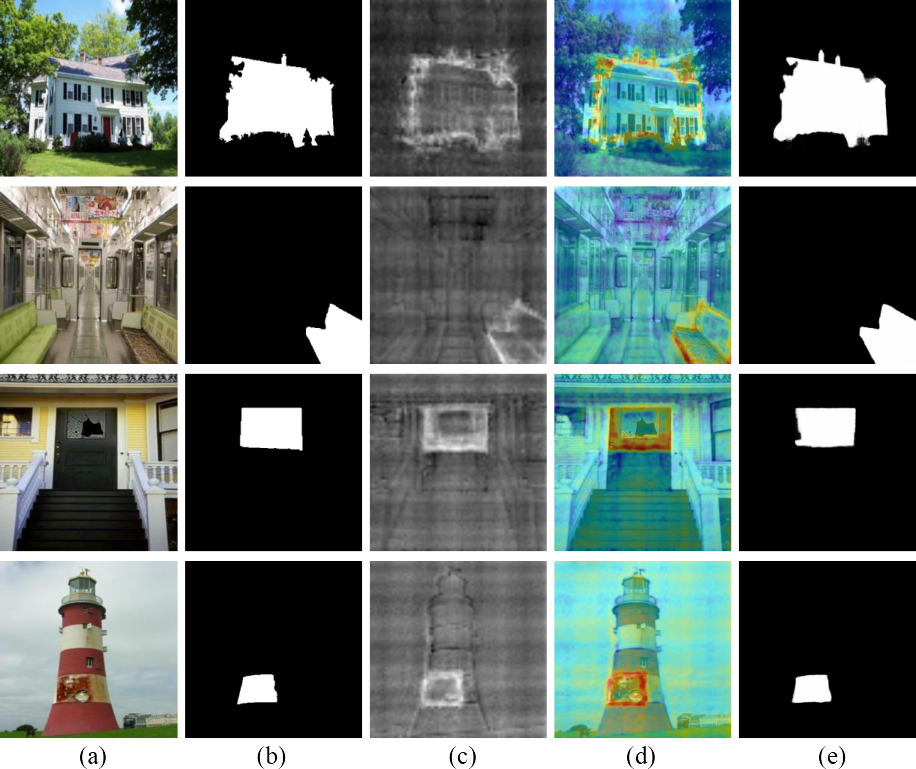}
		\caption{Visualization results of SSGM. (a) Manipulated image; (b) Ground truth; (c) Correction gate map, which takes the mean of the channels and restores them to the 2D dimension; (d) Overlay of gate image and tampered image; (e) Result.}
		\label{fig9}
	\end{figure}
	
	More specifically, SSCR improves refinement by jointly propagating mask and edge cues under multiple scan paths. As illustrated in Fig.~\ref{fig8}, the predicted edge map provides an explicit representation of semantic scope, while the four directional sequence features reveal how the same candidate region is examined under different contextual traversals. A notable observation is that these directional responses are not identical. Instead, they emphasize different structural dependencies around the manipulated region. This is important because a semantic edit is often small and locally plausible. If the model only relies on one propagation path or one type of evidence, it may either miss the decisive edited part or over-activate nearby irrelevant regions. By contrast, SSCR repeatedly verifies the candidate through row-forward, row-backward, column-forward, and column-backward propagation, which helps suppress locally strong but semantically inconsistent responses while preserving those that remain stable across multiple contextual paths. This explains why the full SCR setting produces more compact and semantically coherent predictions than the baseline or the single-branch variants in Fig.~\ref{fig6}.
	
	The edge branch also plays a particularly important role here. In our framework, edge is not introduced merely for conventional boundary sharpening. Instead, it serves as a proxy for semantic scope. In SML, the manipulated content is often meaningful only relative to its surrounding object extent and local structure. Therefore, edge cues help the model estimate where the valid scope of a semantic change lies. This is why using only the mask branch cannot fully recover the final performance. Without explicit scope information, the model may identify the approximate manipulated content but still produce incomplete, expanded, or semantically ambiguous masks. The edge-aware design in SSCR provides an additional constraint that stabilizes the refinement process under content-scope consistency. This behavior is visible in Fig.~\ref{fig8}, where the predicted edge map and the refined result are closely aligned with the semantic extent of the manipulated target.
	
	In addition, SSGM further improves the final prediction by making the contribution of scope cues content-adaptive rather than unconditional. As shown in Fig.~\ref{fig9}, the learned correction gate concentrates on the semantically relevant subregions instead of responding uniformly across the object. This indicates that not all scope information is equally useful for localization. Some structural cues are supportive, while others may be irrelevant or even distracting. By allowing the mask feature to generate a gate that selectively modulates the edge feature, SSGM acts as a semantic relevance filter. The overlay visualization in Fig.~\ref{fig9} further shows that the gate activation is spatially aligned with the manipulated region, and the final result becomes cleaner and more focused after this modulation step. Therefore, SSGM is not a simple fusion block; it is an important mechanism that converts scope cues into selectively useful corrective evidence. Together, SSCR and SSGM make SCR a semantic verification module rather than a conventional mask refinement block, which is exactly why it brings the most critical improvement in our framework.
	
	\subsection{Robust Analysis}
	To further evaluate the robustness of Trace under common post-processing operations, we compare its IoU performance with other methods under Gaussian blur, JPEG compression, and Gaussian noise with different intensities, as shown in Table~IV. Overall, Trace consistently achieves the best performance across all perturbation settings, demonstrating strong robustness to degradations that commonly weaken forensic traces.
	
	A clear trend can be observed under Gaussian blur. As the blur kernel increases from $3$ to $15$, the performance of all methods decreases, since blur suppresses fine-grained structural details and weakens local artifact cues. Nevertheless, Trace maintains a clear advantage over all compared methods across all blur levels. This suggests that our method does not rely solely on fragile low-level traces, but is able to preserve localization ability through semantic anchoring and semantic-constrained reasoning even when local details are smoothed.
	
	A similar phenomenon is observed under JPEG compression. Although compression introduces quantization artifacts and may distort the original manipulation evidence, Trace remains the best-performing method under both tested quality levels. This result indicates that the proposed framework is less sensitive to compression-induced distribution shifts and can still capture semantically meaningful manipulation regions under lossy image reconstruction.
	
	Under Gaussian noise, Trace again achieves the strongest performance across all noise levels. While all methods exhibit gradual degradation as the noise variance increases, the decline of Trace is relatively moderate. This robustness can be attributed to two aspects. First, the semantic anchoring module provides stable object-level priors that are less easily disrupted by noise. Second, the semantic perturbation sensing and semantic-constrained reasoning modules jointly suppress noisy local responses and retain candidate regions that remain semantically coherent under content-scope constraints. Therefore, even when the input is contaminated by strong post-processing perturbations, Trace still maintains reliable localization performance.

	\subsection{Limitation Analysis}
\begin{figure}[t]
		\centering
		\includegraphics[width=0.48\textwidth]{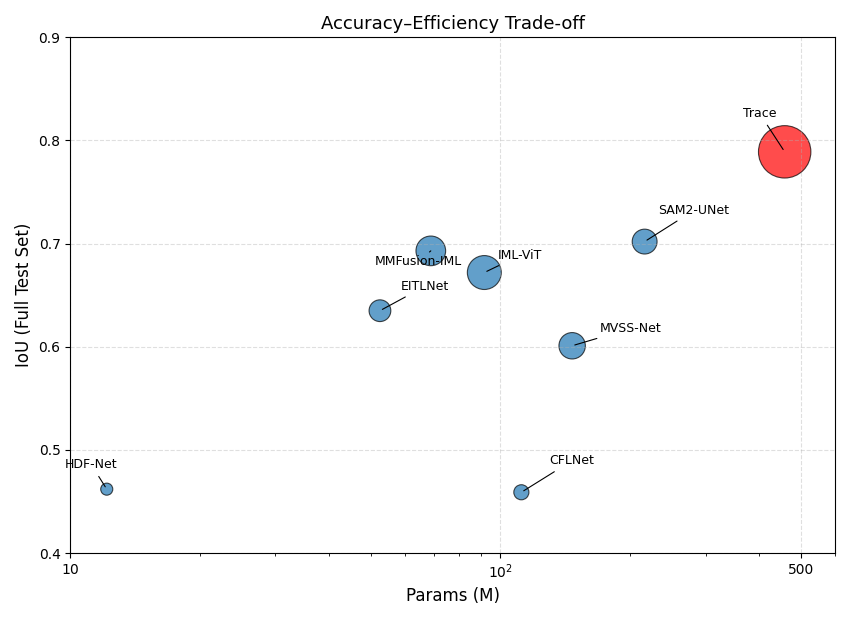}
		\caption{Comparison of accuracy–efficiency trade-offs among different methods. The x-axis denotes model parameters (in millions) in log scale, and the y-axis shows IoU on the full test set. Bubble size encodes computational cost (GFLOPs) with square-root scaling to mitigate the large dynamic range across models. Trace achieves the best IoU, highlighting its effectiveness despite increased computational complexity.}
		\label{fig10}
	\end{figure}

	Although Trace achieves strong localization performance, it still has room for improvement in terms of computational efficiency, which is reported in Fig.~\ref{fig10}. This is mainly because our framework is built upon a large SAM-based backbone and further incorporates frequency-domain adaptation and semantic-constrained reasoning modules. These designs are important for capturing subtle yet semantically meaningful manipulations, but they also introduce additional computation and memory overhead. As a result, the current model is less lightweight than some existing methods and may be less convenient in real-time or resource-limited scenarios. Nevertheless, we view this trade-off as reasonable at the current stage, since our primary goal is to establish a strong baseline for semantic manipulation localization. In future work, we will explore more efficient backbone designs, lighter reasoning strategies, and compact adaptation modules to further improve the balance between performance and efficiency.
	
	\section{Conclusion}
	In this paper, we introduce Semantic Manipulation Localization (SML), a new forensic task that targets the localization of subtle, meaning-altering semantic edits. Different from conventional IML, SML emphasizes semantic sensitivity rather than low-level anomaly detection, and is particularly challenging because manipulated regions are often small, semantically decisive, and visually consistent with surrounding content. To support systematic study of this problem, we construct a dedicated benchmark with pixel-level annotations using a semantics-driven manipulation pipeline. Building on this task, we propose TRACE, an end-to-end framework that integrates semantic anchoring, semantic perturbation sensing, and semantic-constrained reasoning. By grounding localization in semantically meaningful regions, enhancing sensitivity to subtle perturbation cues, and verifying candidate responses under joint content-scope constraints, TRACE achieves clearly superior quantitative and qualitative performance on our benchmark. The results validate the importance of explicitly modeling semantic sensitivity for modern manipulation localization. We hope this work can encourage further research on semantically aware image forensics and inspire more robust localization methods for increasingly realistic editing scenarios.
	
	\bibliographystyle{IEEEtran}
	\bibliography{reference.bib}
	
	\vfill
	
\end{document}